\documentclass[12pt]{article}
\usepackage[margin=2.3cm]{geometry}
\usepackage{graphicx} 

\usepackage[style=apa,backend=biber]{biblatex}
\usepackage{hyperref} 
\usepackage{authblk}
\usepackage{marvosym}
\usepackage[utf8]{inputenc}

\usepackage{xcolor}
\definecolor{refcolor}{rgb}{0.1, 0.1, 0.4}

\hypersetup{
    colorlinks=true,        
    linkcolor=refcolor,     
    citecolor=refcolor,     
    urlcolor=refcolor       
}

\usepackage[most]{tcolorbox}
\tcbset{
  abstractbox/.style={
    colback=gray!5,     
    colframe=gray!5,   
    boxrule=0.3pt,
    arc=1mm,
    left=6pt,
    right=6pt,
    top=4pt,
    bottom=4pt,
    boxsep=5pt
  }
}

\usepackage[font=small,labelfont=bf]{caption}

\usepackage{float}

\usepackage{crimson}

\usepackage{amsmath}

\title{
\textbf{Deep Learning Models Also Recall Features}}
\author[1,2]{Pierre Beckmann}
\affil[1]{École Polytechnique Fédérale de Lausanne (EPFL)}
\affil[2]{Idiap Research Institute}
\affil[ ]{\textcolor{gray}{\small\Letter} \hspace{0.05cm} \href{mailto:pierrebeckmann@gmail.com}{\texttt{pierrebeckmann@gmail.com}}}

\date{February 2026}

\makeatletter
\renewcommand{\maketitle}{\bgroup\setlength{\parindent}{0pt}
\thispagestyle{plain}
\begin{flushleft}
\vspace*{1.1cm} 
  {\LARGE \@title}

  \vspace{0.5cm}

  \@author

    \vspace{0.5cm}

  \@date

  \vspace{0.3cm}

\end{flushleft}\egroup
}
\makeatother

\usepackage{titlesec}

\titleformat{\section}
  {\fontsize{16}{16}\selectfont\bfseries} 
  {\thesection}{1em}{}

\begin{document}

\maketitle
\renewcommand{\thefootnote}{\fnsymbol{footnote}}
\footnotetext[1]{An earlier version of this paper was presented at the 6th Biannual Conference on the Philosophy of Artificial Intelligence (PhAI 2025). I thank participants for helpful feedback, in particular Elliot du Sordet, Iwan Williams, and Kola Ayonrinde.}
\renewcommand{\thefootnote}{\arabic{footnote}}
\setcounter{footnote}{0}

\begin{tcolorbox}[abstractbox]
\noindent
\textbf{Abstract:} 
Recent work in mechanistic interpretability has studied how large language models recall facts stored in their weights. This paper argues that factual recall points to something broader: a general kind of operation in deep learning models, which I call feature recall. The core observation is that a linear projection can be read as retrieving stored information scaled by input activations. I define feature recall, show it applies across architectures, and contrast it with the established paradigm of feature combination. I also consider how cases of feature recall might be mechanistically identified. The account gives philosophers a new conceptual tool for understanding deep learning, and points to empirical directions for mechanistic interpretability research.

\vspace{0.1cm}
\noindent
\textbf{Keywords:} deep learning, feature combination, feature recall, mechanistic interpretability
\end{tcolorbox}

\vspace{0.2cm}

\section*{Introduction}

Recent work in mechanistic interpretability has studied how large language models (LLMs) recall facts \parencite{geva2023dissecting, nanda_fact_2023, chughtai2024summingfactsadditivemechanisms, ameisen_circuit_2025}. These findings point to a way of making sense of internal operations in deep learning models in general, which I call feature recall. It stands in contrast to the well-established paradigm of feature combination.

Feature combination holds that deep neural networks operate by hierarchically combining lower-level features into higher-level ones. This framing is central to the 2015 Nature paper by \citeauthor{lecun2015deep}, who describe deep learning as composing ``simple but non-linear modules that each transform the representation at one level into a representation at a higher, slightly more abstract level." The paradigmatic illustration is image recognition via convolutional neural networks (CNNs), where early layers detect edges, intermediate layers assemble textures, and later layers compose objects \parencite{lecun2015deep, olah2017feature}. This paradigm has been philosophically productive as it has grounded arguments about abstraction in CNNs \parencite{buckner_empiricism_2018, buckner2024deep}, as well as concept formation \parencite{raz_methods_2023} and understanding \parencite{Beckmann2025} in deep learning models.

Yet feature combination does not always offer a satisfying account of what deep learning models do. When ChatGPT produces a biography of Michael Jordan, the biographical content is not contained in the input prompt, waiting to be hierarchically assembled. Instead, the model seems to be retrieving stored information. I argue that such cases are better understood through what I call feature recall. Feature recall would then be a useful complement to the feature combination paradigm, rather than a replacement of it.

I proceed in three steps. First, I explain factual recall in LLMs, a case of feature recall (\S1). I then define feature recall in general terms and show it applies across deep learning architectures (\S2). Finally, I present two readings of feature combination and recall: as interpretive lenses or as operational kinds (\S3).

\section{Factual recall in LLMs}

Factual recall in LLMs is an active area of mechanistic interpretability research \parencite{geva2023dissecting, nanda_fact_2023, chughtai2024summingfactsadditivemechanisms, ameisen_circuit_2025, lindsey_biology_2025}. To present the mechanism, I first introduce some key concepts from the field.\footnote{For a more in depth philosophical introduction to mechanistic interpretability see \textcite{beckmann2026mechanisticindicatorsunderstandinglarge}. For an accessible video introduction to factual recall in LLMs see \textcite{sanderson2024llms}.}

Mechanistic interpretability takes the residual stream as its primary unit of analysis \parencite{elhage2021mathematical}. The residual stream is the evolving internal representation of a token as it passes through the transformer, from initial embedding to final prediction. Rather than viewing each layer as transforming the input into something new, this perspective emphasizes how layers successively add vectors to the evolving token representation.\footnote{This additive interpretation is enabled by the residual/skip connections of the transformer architecture.} The residual stream thus acts as a persistent computational workspace where information accumulates.

The residual stream's vector space is structured by many directions that correspond to internal features, each encoding a particular property that can be present to a varying degree. A helpful metaphor is to think of each feature direction as a slider: as a token flows through the model, attention heads and MLPs write into the residual stream by adjusting the positions of these sliders, setting values that reflect the presence or absence of particular properties.\footnote{The term ``feature" is used both for the direction in latent space (the slider itself) and the activation of a particular input along that direction (the slider value). This dual usage is standard in the field, and I adopt it here.}

\begin{figure}[H]
    \centering
    \includegraphics[width=\linewidth]{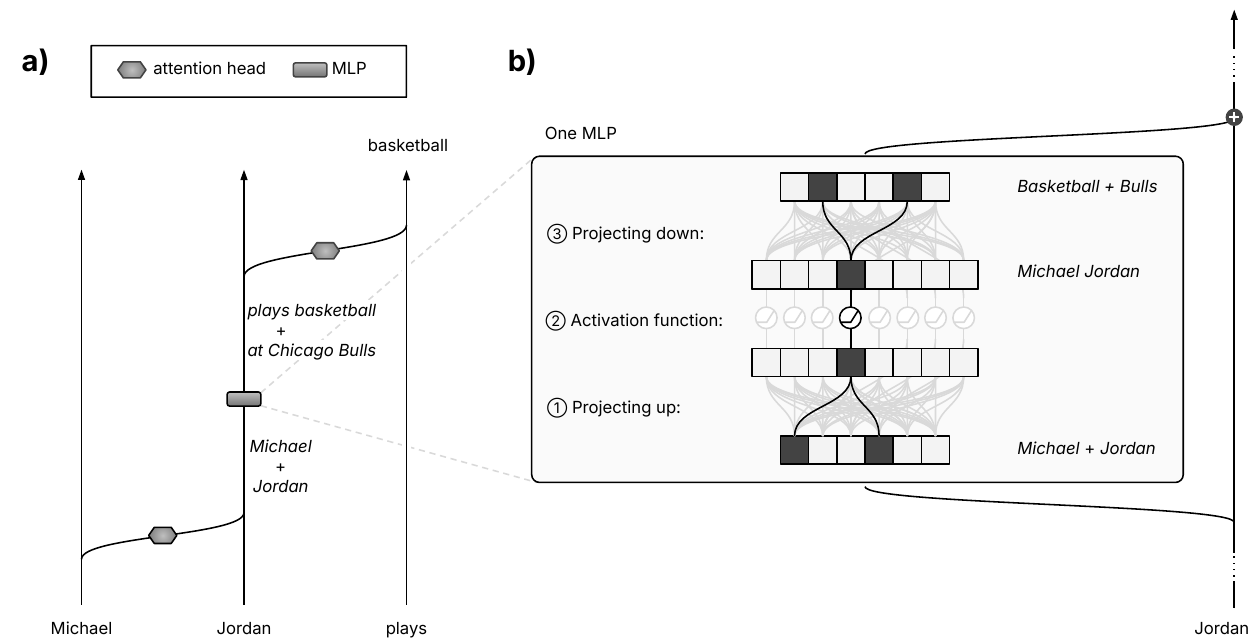}
    \caption[]{An idealized case of feature-activated factual recall. Panel (a) shows the sequential processing of ``Michael Jordan plays," word-token by word-token as they are processed along the residual stream. To complete this sentence (by predicting the ``basketball" token), an attention head first routes the \textit{Michael} feature from the first residual stream to the second. MLP layers then perform two distinct operations. First, they combine the \textit{Michael} and \textit{Jordan} features into a unified \textit{Michael Jordan} feature. Second, they retrieve associated facts stored via the network's weights---such as \textit{plays basketball} or \textit{Chicago Bulls player}---based on this combined feature. Another attention head then routes this retrieved information to the last residual stream, providing the context needed for accurate next-token prediction. Panel (b) details how both feature combination and factual recall could occur within a single MLP's linear transformations in the residual stream of the ``Jordan" token. The upward projection (1) recognizes the pattern of co-occurring \textit{Michael} and \textit{Jordan} features, leading to a strong activation of the \textit{Michael Jordan} feature. The activation function (2) suppresses any spurious activations. Through the downward projection (3) this feature activates the retrieval of certain stored weights, which when added into the output cause the activation of features such as \textit{plays basketball} or \textit{Chicago Bulls player}. The weights connecting the \textit{Michael Jordan} feature to output features directly encode factual associations, ensuring that whenever this feature activates, relevant factual information is automatically retrieved and added to the output representation.\protect\footnotemark}
    \label{fig1}
\end{figure}

The adjustment of these ``slider values" occurs through successive transformer blocks, each containing two key components. Attention heads route information between token positions, enabling the model to integrate context across the sequence. Multi-layer perceptrons (MLPs) perform localized transformations on the current token's representation. Both components operate additively on the residual stream: they write new feature activations into the workspace, thus setting new ``slider values". MLPs' localized transformations encompass both feature combination and what mechanistic interpretability researchers have termed factual recall. To see both operations at work, consider the idealized example in Figure \ref{fig1}, adapted from \textcite{nanda_fact_2023}.

\footnotetext{Some operations such as layer normalization are left out from this idealized case. Furthermore, the figure assumes that each feature corresponds to a single neuron, a property called neuron alignment. I adopt this simplification to make the figure legible and the core mechanism accessible to readers less familiar with the field. Readers familiar with superposition (the finding that models represent far more features than they have neurons, thereby distributing features across neuron directions; \cite{elhage_toy_2022}) may worry that this undermines the mechanism. Thankfully, it doesn't. A feature direction that does not align with a neuron direction still has a well-defined projection through the weight matrix. The matrix maps the input direction to an output vector that can have strong components along multiple output feature directions. The weights then still encode stored associations; the only difference is that these associations hold between directions in activation space rather than between neurons.}

This example shows that once we conceptualize directions in the residual stream as features, linear projections can directly encode associations between them. These associations can represent facts: the \textit{Michael Jordan} $\rightarrow$ \textit{basketball player} connection encodes the fact that Michael Jordan is a basketball player. In its simplest form, such an association can correspond to just a single weight in the model, one value in a projection matrix. One weight can ensure that high activation of the \textit{Michael Jordan} feature will systematically lead to high activation of the \textit{basketball player} feature, creating a direct pathway for retrieval through the network's learned parameters.\footnote{Attention heads can also perform factual recall through their value and output projections (e.g., \cite{geva2023dissecting, chughtai2024summingfactsadditivemechanisms}). But these projections represent only a fraction of attention heads' total parameters, and attention heads collectively contain roughly half as many parameters as MLPs in typical transformer architectures. The majority of in-place computations---including both feature combination and feature recall---occur within MLPs.}

\begin{figure}[H]
    \includegraphics[width=\linewidth]{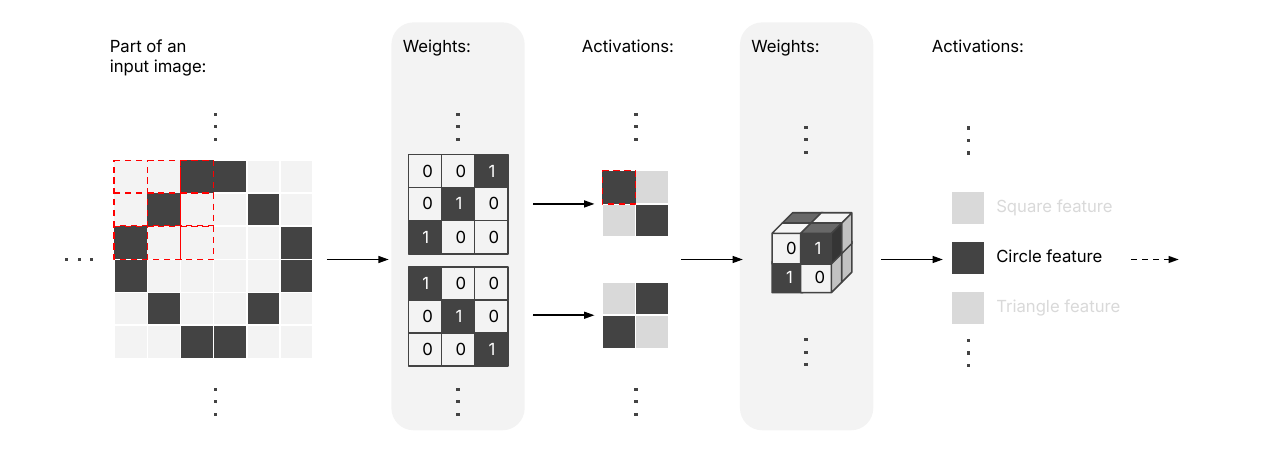}
    \caption[]{Feature combination in a CNN: an idealized case of a circle-recognition circuit. The example traces how a subset of weights of a CNN detects a circle in a 9$\times$9 pixel image region. The process works in two stages. First, the network applies two different 3$\times$3 pattern detectors to non-overlapping regions of the input image. Each kernel scans four 3$\times$3 regions with stride 3, generating two separate 2$\times$2 activation maps. The red box illustrates how this works for the top left 3$\times$3 region and the first kernel: when the top-left kernel perfectly matches its corresponding 3$\times$3 input region, matrix multiplication yields a strong response of 3. The top-right 3$\times$3 region shows only partial overlap with the first kernel at the center pixel, producing a weaker activation of 1. Applying both kernels, creates two 2$\times$2 feature maps. These are then concatenated into a single 2$\times$2$\times$2 tensor (though the stacking isn't shown in the figure). Second, a higher-level 2$\times$2$\times$2 kernel is applied, arranged exactly like our current concatenation of the two feature maps. It will therefore get its strongest activation precisely when the input 9$\times$9 contains a circle. So the network detects the circle hierarchically: from 9$\times$9 pixels, to two 2$\times$2 edge maps, to one circle feature.\protect\footnotemark}
    \label{fig2}
\end{figure}

The key aspect of this interpretation is that the focus shifts from inputs to weights. The input feature merely acts as a trigger; the real work lies in what the weights store. To appreciate how different this is from the feature combination interpretation, consider a CNN that detects circles in images (Figure \ref{fig2}).

\footnotetext{Some key operations of CNNs such as pooling, nonlinear activation functions, and normalization are left out from this idealized case. As with Figure \ref{fig1}, features are shown as neuron-aligned for simplicity.}

It would be odd to say the circuit is retrieving information about circles. It is detecting a pattern by combining lower-level features. The weights encode what configuration of inputs to look for, not what to recall once something is found. That is the core contrast: under the feature combination reading, weights define patterns to detect; according to feature recall, they store information to be retrieved.

Factual recall names one instance of a broader mechanism. I propose to call the general operation feature recall, for two reasons. First, feature recall need not always rely on factual associations; it can also encode conceptual ones. The \textit{Michael Jordan} $\rightarrow$ \textit{basketball player} connection is factual: a contingent, empirical association that could have been otherwise. But when a ``bachelor" feature triggers ``unmarried," the association is conceptual, not factual, as being unmarried is part of what it is to be a bachelor. Second, factual recall makes it sound as if the mechanism always recalls ``correct" facts. But feature recall relies on putative connections the model formed during training, and these can be wrong. Feature recall is thus the more general and more precise term. It names the mechanism without assuming what kind of connection is encoded or whether it is correct.

Having presented a precise case of feature recall in LLMs, I now generalize the observation. I argue that feature recall is not specific to one architecture but a general operation that all deep learning models can be seen as performing.

\section{Feature recall in deep learning models}

At its core, every major deep learning architecture relies on matrix multiplication between an input vector and a learned weight matrix, with the addition of fixed layers such as activation functions. This notably holds for convolutional kernels in CNNs, weight matrices in MLPs, value and output projections in attention heads, and gate matrices in LSTMs. These matrix multiplications are constrained differently by the architecture but essentially always define a projection between an input and an output layer.

\begin{figure}[h!]
    \centering
    \includegraphics[width=\linewidth]{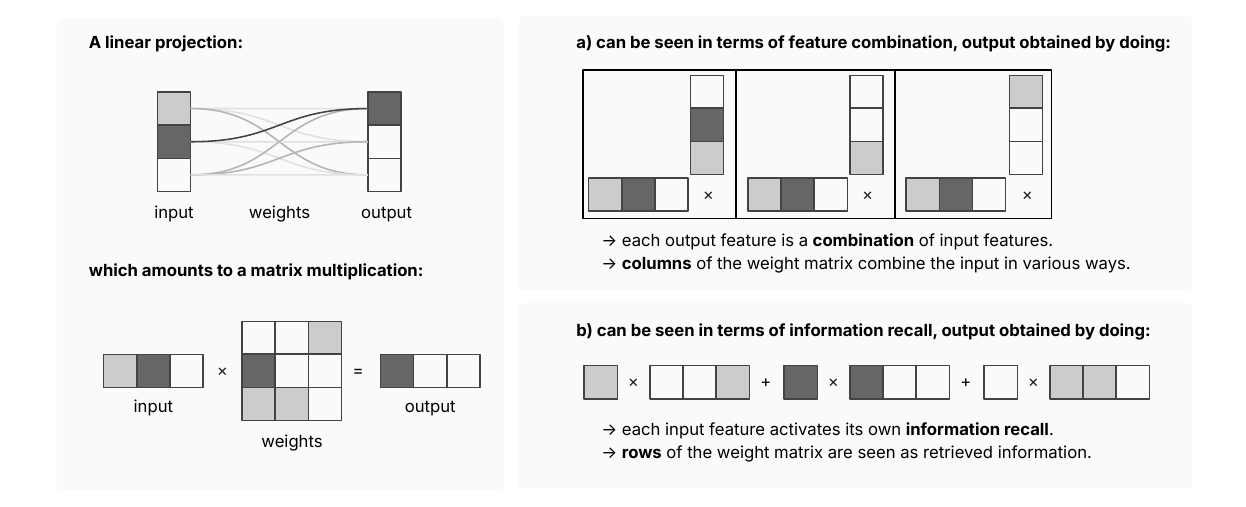}
    \caption{A linear projection, or matrix multiplication, can be seen both in terms of (a) feature combination or (b) feature recall. In case (a), each output element is obtained by multiplying the input vector with a column of the weight matrix. This can be seen as feature combination: each column defines a pattern of how input features should be combined to obtain one specific output feature. In case (b), each input feature multiplies out one row of the weight matrix and the resulting vectors are summed (note that this amounts to the exact same operation because matrix multiplication can be decomposed as either a dot product with columns or a weighted sum of rows). Each row can then be seen as a stored set of potential feature activations that gets scaled by the input features---the operation is thus seen in terms of feature recall.
}
    \label{fig3}
\end{figure}

The factual recall mechanism from Section 1 suggests a new way to read any such projection. Consider a simple MLP layer that transforms an input vector of size 3 into an output vector of size 3 (Figure \ref{fig3}).

Feature combination is many-to-one: many input features contribute to producing one output feature, and the weights specify what to detect. Feature recall, by contrast, is one-to-one or one-to-many: one input feature triggers the retrieval of one or many output features, and the weights specify what to retrieve. The input acts as a trigger; the substantive content of the transformation resides in what the weights store. This is exactly what we saw in the downward projection of the MLP in Figure \ref{fig1}b, with the \textit{Michael Jordan} feature scaling out a row of stored associations, retrieving \textit{basketball player}, \textit{Chicago Bulls}, and so on.

Since every deep learning architecture relies on learned linear projections, feature recall applies wherever deep learning operates. More precisely, feature recall is the operation by which input feature activations scale stored weight vectors, thereby retrieving associated output feature activations. To illustrate, return to the circle-detecting CNN from Figure \ref{fig2}, but suppose now that it is part of a model trained to classify handwritten digits in MNIST. Once the circle feature activates, it could scale out a row of stored weights, retrieving features for digits that typically contain circles, such as 0, 6, 8, and 9, and feeding these into downstream processing. The model would thus be recalling which digits contain circles to help in its classification task. So even in a CNN where a lot of feature combination happens, some projections might be best understood through the lens of feature recall.

More generally, feature recall helps make sense of cases that were puzzling under a pure feature combination reading. When a language model generates a detailed biography of Michael Jordan from a short prompt, the biographical content is not present in the input. Feature recall offers a natural explanation: stored associations are retrieved and chained through successive projections, letting a few words of input yield a long, detailed answer. The same logic applies to image and video generation models, which can produce complex outputs from just a few words of prompt.

Finally, I take it that feature recall opens the way to thinking about dispositional beliefs in deep learning models. So far, discussions of belief in LLMs have focused on occurrent beliefs, that is, beliefs that are active during a particular inference and therefore identifiable through linear probes on residual stream activations (see \cite{herrmann2025standards}). But the weight-based associations underlying feature recall have a distinctly dispositional character. The fact that Michael Jordan plays basketball is encoded in the model's weights regardless of whether the model is currently processing anything about Michael Jordan. It is there as a standing disposition, ready to be triggered when the relevant feature activates. This suggests a natural mapping: occurrent beliefs correspond to features encoded in activations at a given inference, while dispositional beliefs correspond to associations stored in weights, retrieved only when the right trigger arises. I leave the careful exploration of this connection for future work, but if the mapping holds, feature recall could provide a framework for locating dispositional belief in deep learning models, a category that has so far received little attention.\footnote{I thank Iwan Williams for suggesting this connection.}

\section{Interpretive lens or operational kind?}

Throughout this paper, I have treated feature combination and feature recall primarily as interpretive lenses: two ways of reading the linear projections of deep learning models. A careful reader may have noticed something troubling about this. As Figure \ref{fig3} illustrates, any matrix multiplication can be decomposed either as dot products with columns (feature combination) or as a weighted sum of rows (feature recall). The two readings are mathematically equivalent, which threatens to collapse the distinction into a mere difference in perspective. What should be done about this?

One option is to bite the bullet and accept that the distinction is ultimately hermeneutic. Feature combination and feature recall are ways for us to make sense of what deep learning models do, with different projections lending themselves more naturally to one reading or the other. On this view, the circle-detecting kernel is most naturally read as feature combination, while the MLP projection retrieving facts about Michael Jordan is most naturally read as feature recall. But there is no fact of the matter beyond what we find explanatorily useful. This comes at a cost, however: it blurs the line between the two operations, since recalling many features from one input could, in a sense, be redescribed as combining one input with many stored associations (and vice versa).

The more ambitious option is to try to draw a hard line between the two, so that any given learned projection falls cleanly into one category or the other. Since the mathematical formalism alone cannot distinguish the two (any projection admits both readings), this requires anchoring the concepts to an empirical criterion that forces them apart. Such a criterion would identify a property of trained projections that makes one reading apt and the other inapt for a given case. I sketch here what such a criterion might look like. The proposal may be of use to the mechanistic interpretability researcher (see \cite{williams2025}), and it illustrates to the philosopher how a conceptual distinction can be
operationalized for deep learning.

The structural asymmetry between the two operations provides a natural starting point. Feature combination is many-to-one, with many input features contributing to one output feature; feature recall is one-to-one or one-to-many, with one input feature dominating the output. This amounts to a difference in effective connectivity: feature combination draws on a dense set of input features, while feature recall routes through a sparse one, often just a single input triggering the retrieval of stored associations. To capture this difference quantitatively, one could define a connectivity ratio such as:

\begin{equation*}
CR_k = \frac{\max_i |w_{ik}|}{\sum_i |w_{ik}|}
\end{equation*}

\noindent
which, for a given output feature $k$ in a projection with $n$ input features, measures how concentrated the weight mass is on a single input. A high ratio (closer to 1) indicates sparse effective connectivity, where one input does most of the work---suggesting that feature $k$ was produced by feature recall. A low ratio (closer to 0) indicates dense effective connectivity, where many inputs jointly shaped the output---suggesting that feature $k$ was produced by feature combination.

One important caveat is that the connectivity ratio should be computed over projections connecting features, not raw neuron activations. Individual neurons rarely correspond to single features; instead, multiple features are typically superposed within the same neurons \parencite{elhage_toy_2022}. A preliminary decomposition step, using dictionary learning methods such as transcoders \parencite{bricken_towards_2023, templeton_scaling_2024, ameisen_circuit_2025}, would be needed to recover interpretable features before the ratio can be meaningfully applied.

Whether this criterion carves at useful joints is an empirical question. I can think of two lines of inquiry to test here. The first would apply the connectivity ratio to a trained model and examine whether projections with sparse connectivity correspond to recognizable cases of retrieval, such as factual or conceptual associations. The second would compute the ratio across many output features and examine the shape of the resulting distribution. A bimodal distribution would suggest that the criterion reveals a precise boundary between the two concepts.\footnote{A further direction would be to define a context-dependent variant of the ratio, for example by scaling weights by actual input activations on a given forward pass. This might allow identifying whether a particular projection is performing feature recall or feature combination in a specific context, rather than in general.}

If the criterion works, it could open several avenues of investigation, such as how the balance between feature recall and feature combination varies across model sizes, architectures, or layers. To take one concrete case: \textcite{gupta2025llmsusedepth} describe a ``guess-then-refine" pattern in which early transformer layers produce many guesses about what token might come next, while later layers sort through these candidates to select the best prediction. If the criterion tracks a real distinction, one would expect it to pick this up as more feature recall in early-layer MLPs and more feature combination in later ones.

In sum, the question of how to cash out the distinction admits two answers. On a weaker reading, it is purely interpretive, and already useful as such. On a stronger reading, the distinction might become operational where some projections count as one and not the other, given an appropriate mechanistic criterion. Pursuing this stronger reading would require the kind of empirical investigation sketched here.

\section*{Conclusion}

Starting from recent findings concerning factual recall in LLMs, this paper introduced feature recall as a complement to the established paradigm of feature combination in deep learning. Where feature combination describes how models detect patterns to identify higher-level features, feature recall captures how input features trigger the retrieval of information stored in learned weights. The insight is grounded in the simple observation that any linear projection can be read not only as combining inputs to produce outputs, but also as retrieving stored associations scaled by input activations. Feature recall makes intuitive sense of operations that feature combination struggles to explain, most notably how models generate rich outputs from sparse inputs. It should be useful to philosophers working on the attribution of mental states and capacities to LLMs---for instance, the dispositional beliefs discussed above. Whether a mechanistic criterion can sharply separate feature recall from feature combination, elevating the distinction from a useful interpretive heuristic to a principled operational one, remains an open question.

\clearpage

\printbibliography

\end{document}